\documentclass[letterpaper]{article} 
\usepackage{aaai23}  
\usepackage{times}  
\usepackage{helvet}  
\usepackage{courier}  
\usepackage[hyphens]{url}  
\usepackage{graphicx} 
\usepackage{natbib}  
\usepackage{caption} 
\usepackage{algorithm}
\usepackage{subfigure}
\usepackage{newfloat}
\usepackage{listings}
\DeclareCaptionStyle{ruled}{labelfont=normalfont,labelsep=colon,strut=off} 
\floatstyle{ruled}
\newfloat{listing}{tb}{lst}{}
\floatname{listing}{Listing}
\title{MobilePTX: Sparse Coding for Pneumothorax Detection Given Limited Training Examples}
\author{
    Darryl Hannan\textsuperscript{\rm 1},
    Steven C. Nesbit\textsuperscript{\rm 1},
    Ximing Wen\textsuperscript{\rm 1},
    Glen Smith\textsuperscript{\rm 1},
    Qiao Zhang\textsuperscript{\rm 1},
    Alberto Goffi\textsuperscript{\rm 2},
    Vincent Chan\textsuperscript{\rm 2},
    Michael J. Morris\textsuperscript{\rm 3},
    John C. Hunninghake\textsuperscript{\rm 3},
    Nicholas E. Villalobos\textsuperscript{\rm 3},
    Edward Kim\textsuperscript{\rm 1},
    Rosina O. Weber\textsuperscript{\rm 1}
    and Christopher J. MacLellan\textsuperscript{\rm 4}
}
\affiliations{
    \textsuperscript{\rm 1}Drexel University, Philadelphia, PA, USA\\
    \textsuperscript{\rm 2}University of Toronto, Toronto, ON, CA\\
    \textsuperscript{\rm 3}Brooke Army Medical Center, Fort Sam Houston, TX, USA\\
    \textsuperscript{\rm 4}Georgia Institute of Technology, Atlanta, GA, USA\\
    
    dwh48@drexel.edu,
    cmaclell@gatech.edu
}

\begin{document}

\maketitle

\begin{abstract}
    Point-of-Care Ultrasound (POCUS) refers to clinician-performed and interpreted ultrasonography at the patient's bedside. Interpreting these images requires a high level of expertise, which may not be available during emergencies. In this paper, we support POCUS by developing classifiers that can aid medical professionals by diagnosing whether or not a patient has pneumothorax. We decomposed the task into multiple steps, using YOLOv4 to extract relevant regions of the video and a 3D sparse coding model to represent video features. Given the difficulty in acquiring positive training videos, we trained a small-data classifier with a maximum of 15 positive and 32 negative examples. To counteract this limitation, we leveraged subject matter expert (SME) knowledge to limit the hypothesis space, thus reducing the cost of data collection. We present results using two lung ultrasound datasets and demonstrate that our model is capable of achieving performance on par with SMEs in pneumothorax identification. We then developed an iOS application that runs our full system in less than 4 seconds on an iPad Pro, and less than 8 seconds on an iPhone 13 Pro, labeling key regions in the lung sonogram to provide interpretable diagnoses.
\end{abstract}

\section{Introduction}

Ultrasound imaging techniques are crucial to many medical procedures and examinations. The development of portable ultrasound devices has allowed healthcare professionals to perform and interpret sonographic examinations with the goal of making immediate patient care decisions wherever a patient is being treated, including out-of-hospital scenarios. This clinician-performed and interpreted ultrasonography at the patient's bedside has been referred to as Point-of-Care Ultrasound (POCUS). While POCUS allows ultrasound to be used in a variety of new tasks and settings, a few bottlenecks still remain. For example, both the acquisition and interpretation of sonograms requires specific training and competency development. Moreover, the quality of the ultrasound device and the proficiency of the operator collecting the data can lead to challenging cases that are uninterpretable or elicit disagreement among experts.

The development of AI systems that can assist medical professionals in interpreting sonograms by highlighting key regions of interest (ROIs) and suggesting potential diagnoses would increase both the accuracy and the efficiency of differential diagnosis. However, developing intelligent systems that are capable of operating in the medical domain is challenging due to the lack of annotated data. Many computer vision applications rely upon thousands, or even millions, of examples to achieve reasonable performance. In the medical domain, collecting labels is an expensive process because a high degree of expertise is required to interpret the images. Additionally, data collection for some tasks can be challenging due to some procedures only being performed in specific high-stakes situations (e.g., the patient's life might be at risk and immediate action is required). In the domain of real-world images, transfer learning techniques are frequently used to compensate for a lack of labeled data, however they are suboptimal for the medical domain due to a lack of visual feature overlap.

\begin{figure}[t]
    \centering
    \includegraphics[width=0.85\columnwidth]{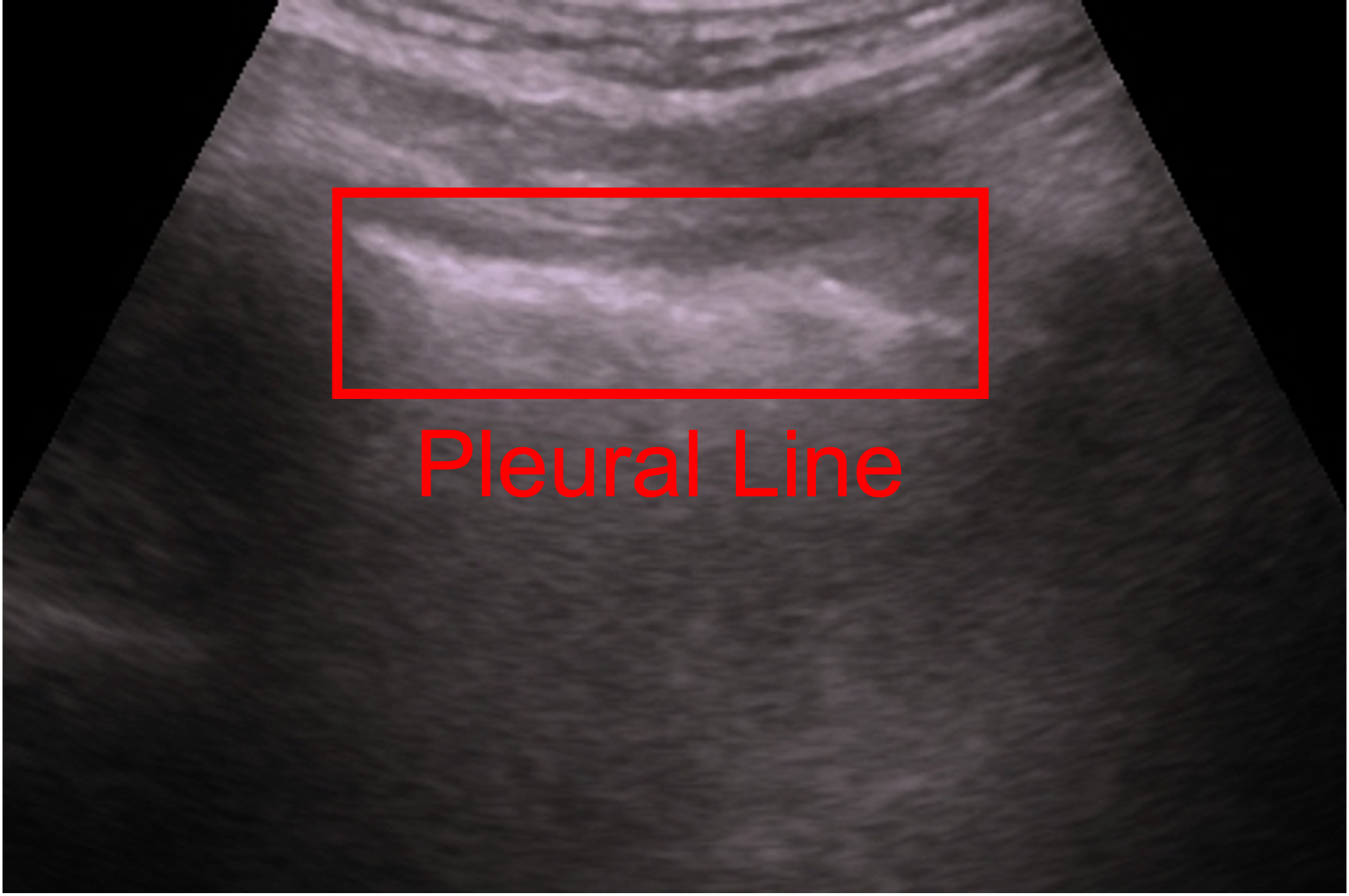}
    \caption{Sample lung sonogram from the COVID-19 dataset, highlighting the pleural line.}
    \label{fig:ultrasound_ex}
\end{figure}
In this paper, we propose a POCUS mobile application within the context of an ultrasound video classification task. The application must classify a lung ultrasound (LUS) video according to the presence of pneumothorax (PTX) symptoms. Before constructing the system, we sought to gain insight into the process that experts use to make their diagnoses and inject this information into our architecture. \citet{smith2022doctor} conducted a think-aloud analysis where two subject matter experts (SMEs) analyzed and diagnosed videos of patients that potentially had PTX. Their study determined that the pleural line (where the lung comes into contact with the chest wall) and its movement were the most important features for PTX diagnosis. Clinicians can observe two types of pleural line movement in a normal lung: lung sliding (shimmering movement synchronous with respiration) and lung pulse (rhythmic movement of the pleural line at the cardiac frequency). The pleural line remains relatively stationary in a patient with PTX, displaying no sliding and no pulse. Figure \ref{fig:ultrasound_ex} shows a still image from an LUS with the pleural line highlighted.

We leveraged this expert knowledge by decomposing the task into three stages. The first stage extracts the pleural line from the video, then a second stage uses a 3D sparse coding model to extract a sparse representation of the pleural line clip \cite{paiton2019analysis}. The sparse coding model contains biologically inspired mechanisms, such as lateral inhibition, that result in high quality representations with orthogonal features. This model also operates over multiple frames, capturing movement in the video. The final stage passes the sparse representation to a small convolutional neural network (CNN) classifier.

Due to the difficulties surrounding collection and labeling of medical images, we challenged ourselves to minimize the amount of labeled data used to develop our PTX application. In our primary benchmark we used just 47 LUS videos, each approximately 3 seconds long, and trained our model from scratch. To complement the portability of POCUS devices, we ensured that our mobile application could execute on portable, relatively inexpensive hardware. We selected a 12.9-inch Apple iPad Pro as our target device for our experiments and developed our application for iOS 15.

We demonstrate that our application is capable of performing on par with SMEs and that it outperforms a comparable VAE-based architecture and Mini-COVIDNet \cite{awasthi2021mini} on PTX and COVID-19 datasets. Then, we analyzed the impact of further restricting training data, illustrating our model's robustness to learning from limited examples, and evaluated our model in a transfer learning setting, where features are trained on one task and applied to another, demonstrating the importance of our sparse filters. We also qualitatively analyzed our learned filters and discuss efforts aimed at interpretability, and provide an overview of our mobile app and various challenges that we need to address for our application to be deployed.

In summary, we make the following contributions:
\begin{enumerate}
    \item We propose an LUS video classification framework based upon features that experts identified as important for diagnosing PTX.
    \item We demonstrate that the framework outperforms existing architectures and achieves accuracy on par with SMEs, despite only being exposed to a few dozen labeled videos.
    \item We construct an iOS application that executes our model in just a few seconds, illustrating that the framework can be deployed in clinical settings at low cost.
\end{enumerate}

\section{Related Work}
Due to the COVID-19 pandemic, LUS has received heightened attention by the machine learning community. There are two commonly used public datasets targeting this task, ICLUS-DB \cite{roy2020deep} and the COVID-19 dataset \cite{born2021accelerating}. The most common LUS task formulation consists of frame classification and semantic segmentation, with some papers including video classification as well. Many works rely upon a CNN-based architecture for feature extraction \cite{diaz2021deep}, then aggregate frame scores for a video-level prediction \cite{mento2021deep,roy2020deep}. Some works extend this by adding a temporal component to the model, allowing it to detect changes in the video over time \cite{barros2021pulmonary,lum2021imaging,dastider2021integrated}. \citet{ebadi2021automated} opted for a 3D CNN with a separate optical flow branch for capturing these dependencies. Some work has sought to explicitly leverage expert annotations to improve performance. \citet{frank2022} used a convolutional network with not only the B-mode frame as input but also a vertical artifact mask and pleural line heatmap. \citet{chen2021quantitative} took an approach more similar to our own, where instead of classifying the whole video, the model only classifies an ROI centered on the pleural line. They determined the ROI by automatically selecting the highest peak in a radon transformation, whereas in our work we used a separately trained object detection model.

Our work is more application-driven than many of these works; we learned with limited data and ran our model on a mobile device. For these reasons, \citet{awasthi2021mini} is most similar to our work. The researchers trained a frame-based classification model based upon MobileNet \cite{howard2017mobilenets} on just 1,103 ultrasound images. They then demonstrated that their model was able to run on two different embedded systems. One major difference between their work and our own is that we focused on video-level classification and handled temporal information by processing video clips instead of individual image frames. This places a larger burden on the model, making it more difficult to execute on a mobile device. Another difference is that we leveraged expert knowledge expressed in a think-aloud analysis \cite{smith2022doctor} to construct our pipeline. This leads to a greater degree of interpretability in the form of an ROI bounding box around the pleural line.

\section{Application Context}

The primary task that we focus on is PTX (i.e., abnormal collection of air in the pleural space between the lung and chest wall, potentially causing significant lung collapse) detection. The requirements of this task specify that an agent must analyze a LUS video and determine whether the patient exhibits the signs of PTX (positive) or not (negative). The primary artifact of interest is the pleural line, which takes the form of a bright horizontal white line that extends across part of the video (Figure \ref{fig:ultrasound_ex}). In a healthy patient, one can see the visceral and parietal pleura sliding relative to each other at the pleural line during respiration, as well as pulsing in synchrony with cardiac contraction. In the case of PTX, air accumulation between the visceral and parietal pleura separates these membranes and eliminates the sliding motion between them, as well as the cardiophasic movement at the parietal pleura.\footnote{There are some cases where no sliding/no pulse might be observed, yet the patient does not have PTX. In this work we assume that all cases that exhibit these features are indicative of PTX.} We focused our model on these pleural line characteristics based on the think-aloud analysis by \citet{smith2022doctor}.

We considered several design principles when constructing our model, including the type of equipment that would be needed to process videos in a clinical setting, the amount of labeled data available to train the model, and the time required to generate results. Many ultrasound devices support iOS; therefore we developed our mobile application for this platform and explored using both an Apple iPad Pro and an iPhone 13 Pro. We also required our system to produce a prediction in less than 5 seconds after the conclusion of the video.\footnote{These requirements come from the DARPA POCUS program.}
Beyond these application considerations, we also focused on making model training inexpensive by requiring as little annotated data as possible. Due to the high cost of data acquisition and processing, we wanted to add the additional constraint of restricting dataset size, demonstrating that we can still obtain strong results with a relatively small amount of high-quality data labeling. Therefore, we used just 47 annotated LUS videos to train our system for our primary task.

\section{Model Description}

The model that we developed consists of three primary components: a YOLOv4 object detection model \cite{bochkovskiy2020yolov4} that locates the pleural line, a 3D convolutional sparse coding model \cite{paiton2019analysis} that extracts meaningful representations and compresses temporal information, and a classifier that produces a binary classification indicating if the pleural line clip displays movement.

\begin{figure*}[t]
    \centering
    \includegraphics[width=0.9\textwidth]{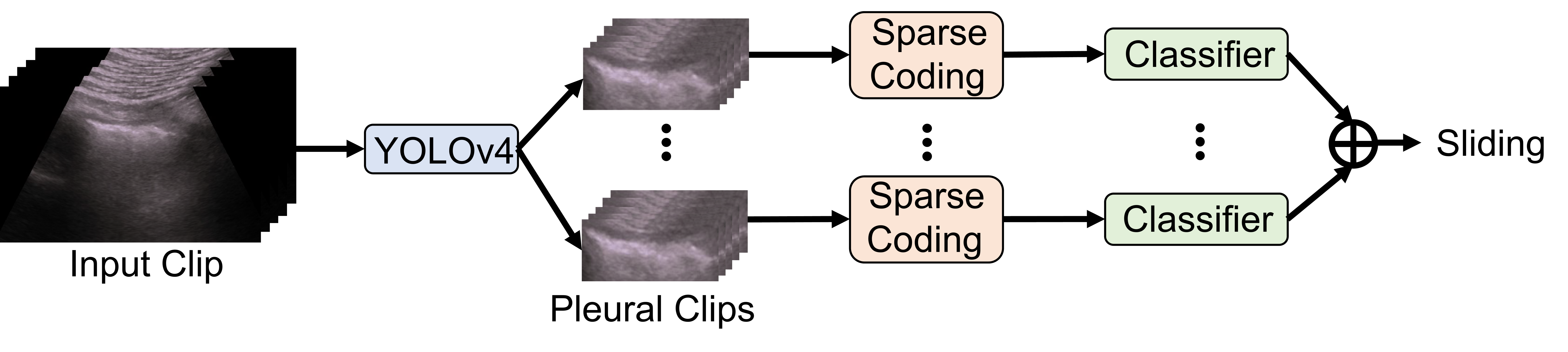}
    \caption{Overview of our PTX classification architecture.}
    \label{fig:model}
\end{figure*}

Processing the entire video using this pipeline would result in our application exceeding our allotted execution time. Therefore, we developed a voting strategy where the model individually processes a number of clips, then aggregates the predictions to produce the final overall video prediction. Our application extracts clips by striding over the video frames at a fixed interval. At each point, the model extracts the given frame along with the two previous and two subsequent frames, resulting in a five-frame clip.
The YOLOv4 model processes the middle frame of this five-frame clip and places a bounding box around the pleural line. Our application applies this same box across all frames in the five-frame clip, producing a final clip that only contains the region around the pleural line. The sparse coding model then encodes this clip and creates a sparse representation, which it sends to the classifier to make a clip-level prediction. To get the final video prediction, the system takes the mode of the clip predictions. In the case of a tie (e.g., two clips are predicted as movement and two as no movement) the application considers the output logits as confidence values, averaging these and rounding to the closest prediction.

\paragraph{Sparse Coding}

One way of minimizing the amount of labeled data required to train a model is to use unsupervised training techniques. Therefore, we leveraged 3D convolutional sparse coding \cite{olshausen1997sparse,paiton2019analysis} to create a sparse representation of our pleural line clips. Sparse coding relies upon biologically plausible learning techniques to learn a dictionary of convolutional filters.

We used a convolutional variant of the Locally Competitive Algorithm (LCA) to compute our sparse features, following \citet{paiton2019analysis} as a guide while developing our models. One can think of this model as an autoencoder that seeks to learn a set of filters, or dictionary elements, $\Phi$ that accurately reconstruct an input video $x$. The encoder produces an activation map $a$, which the decoder then deconvolves with the filters to produce the reconstruction $\hat{x}$.

One can consider the encoder as a recurrent network where an internal state, or membrane potential, $\mu$ is updated at each timestep. The model trains via gradient descent to minimize the energy function:

\begin{equation} \label{eq:energy}
    E(t) = \frac{1}{2}\sum_{i=1}^N[x_i-a(t)\Phi]^2 + \lambda|a(t)|
\end{equation}

where the first term is the sum squared error between the input and reconstruction and the second term is the sparsity penalty. The activation map $a(t)$ is the thresholded internal state $\mu(t)$, where values less than $\lambda$ are set to 0.

\citet{paiton2019analysis} shows that iteratively updating $\mu$ according to Equation \ref{eq:sparse_update} (where $\eta$ is learning rate) minimizes the energy function in the convolutional case.

\begin{equation}
    \mu(t+1) = \mu(t) + \eta (e\ast\Phi + a(t) - \mu(t))
    \label{eq:sparse_update}
\end{equation}

As shown in Equation \ref{eq:sparse_update}, convolution ($\ast$ operation) of the reconstruction error over the filters drives the membrane update. The number of timesteps over which $\mu$ updates is a hyperparameter that one adjusts through observation of when the objective function converges. During the computation of the activation map, the filters do not update. After the model produces the final activation map, the filters then unfreeze. The training algorithm then uses Equation \ref{eq:energy} to compute the final loss and updates the filters using gradient descent.

Previous research has shown that sparse coding can produce robust, semantically meaningful visual features across a variety of tasks, from learning face classifiers \cite{kim2018deep} to aligning binocular video \cite{lundquist2017sparse}. It is even robust to adversarial attacks \cite{schwartz2020regularization}. However, due to its recurrent nature, sparse coding has a higher computational cost than a standard CNN.

\paragraph{Small Convolutional Classifier}

Utilization of YOLOv4 and sparse coding allowed our classifier to have a minimal number of parameters. The model first maxpools the sparse coding activation map. Then it passes the resulting representation through a CNN consisting of 2 convolutional layers and 2 feed-forward layers with dropout. We trained the classifier with a binary cross-entropy loss function.

\section{Experiments}

 We first demonstrate that our model outperforms both a VAE baseline and Mini-COVIDNet \cite{awasthi2021mini} on both our PTX task and an auxiliary COVID-19 LUS classification task. We constructed the VAE baseline to have approximately the same architecture as our model, replacing 3D sparse coding with standard 3D convolution. This also functions as an ablation of our sparse coding model, demonstrating its utility. We additionally present results with Mini-COVIDNet to compare to the most closely related work. We report both accuracy and Macro F1 score to account for imbalanced classes. We performed all evaluations using patient grouping, where splits are such that the same patient never appeared in both our training and test sets.

We performed an analysis in which we reduced the amount of training data by fixed intervals, demonstrating that our model is more robust to training data reduction than Mini-COVIDNet. We then explored the benefits of sparse coding by evaluating our model using sparse filters that were learned from a related LUS dataset. This is a form of transfer learning in which a set of weights learned on one task are applied to another task, typically with some slight adjustment of the weights or additional layers that are learned on top of the transferred network. These filters improved our PTX accuracy compared to randomly initialized filters, illustrating the benefits of learning sparse features on related data.

\paragraph{Model Implementation Details}
We developed our models in Keras and Tensorflow and exported our model to TFLite to run in our iOS app. For sparse coding, we updated the membrane potentials using Adam \cite{kingma2015adam} and used SGD for the filter updates. We trained with a batch size of 32, a filter learning rate of 3e-3, 48 filters of width 15, height 15, and depth 5, a stride of 1, 300 inner loop updates, a membrane potential learning rate of 0.01, a lambda of 0.05, an input clip height of 100 and width of 200, for 100 epochs. We trained the classifier using Adam and reduced the computational cost of sparse coding by using a stride of 2 and 150 inner loop updates with our learned filters. We trained the classifier with a learning rate of 5e-4 for 25 epochs. The VAE shares many of the same hyperparameters, with the exception of 32 filters for 3D convolution, a learning rate of 5e-4, and 40 epochs of training. We trained all models on a single Nvidia A40. We converted videos to grayscale and normalized to a mean of 0. We randomly rotated each input by up to $\pm45^{\circ}$ and randomly applied horizontal flips.

\begin{table}[]
    \centering
    \begin{tabular}{|c|c|c|}
        \hline
        Model & Macro F1 & Accuracy \\\hline
        VAE + Dense & 45.2 $\pm$ 5.8 & 57\% \\
        Mini-COVIDNet (15 pos) & 57.7 $\pm$ 15.1 & 60\% \\
        Mini-COVIDNet (30 pos) & 70.2 $\pm$ 5.5 & 71\% \\
        Our Model & \textbf{87.8 $\pm$ 4.3} & \textbf{88\%} \\\hline
        Subject Matter Experts & - & 91\% \\\hline
    \end{tabular}
    \caption{F1 scores (over 5 runs) for the BAMC dataset.}
    \label{tab:bamc_results}
\end{table}

\subsection{Pneumothorax}
Our primary dataset, the Brooke Army Medical Center (BAMC) dataset, consists of 62 LUS videos, 30 from patients with PTX (i.e., pleural line movement absent) and 32 from patients without PTX (i.e., pleural line movement present). Physicians diagnosed PTX radiographically and 3 expert reviewers confirmed the presence of lung sliding. The average length of these videos is approximately 3 seconds at 20 frames per second. BAMC collected all LUS videos with a convex probe at depths ranging from 4--12 cm. For our sparse coding model, we utilized all 62 videos, however the algorithm is unsupervised so we did not use labeled data for this stage of training. For training our classifier, DARPA challenged us to only use 15 labeled `No movement' videos.

\begin{figure}[t]
    \centering
    \includegraphics[width=0.99\columnwidth]{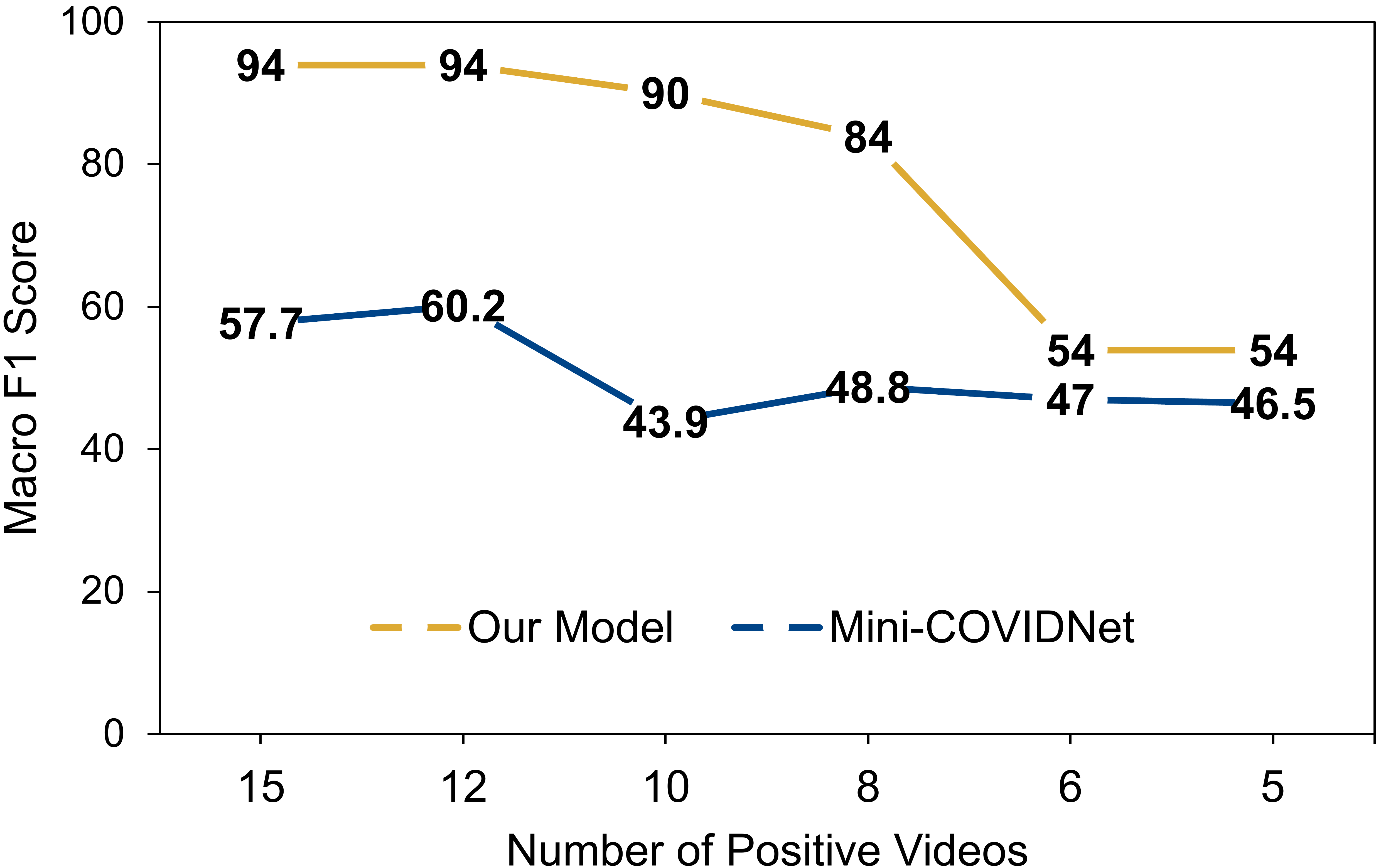}
    \caption{Macro F1 score l with varying quantities of positive training examples.}
    \label{fig:less_data}
\end{figure}

Table \ref{tab:bamc_results} shows the results of our BAMC PTX experiments. Our model achieved an F1 score of 87.8 on our test set and it outperformed both our VAE baseline and Mini-COVIDNet, which obtained F1 scores of 45.2 and 70.2 (30 pos), respectively. We exposed our sparse coding model to all 30 positive examples (without labels), yet exposed our classifier to only 15 positive examples (with labels). Therefore, we evaluated Mini-COVIDNet in both a 15-positive labeled examples setting, where it had a slight disadvantage, and a 30-positive labeled examples setting where it had a clear advantage. Two SMEs evaluated the test set and achieved 91.1\% agreement, slightly outperforming our model. We attribute our model's success to its emphasis on learning robust visual features despite having limited data. We conducted a series of experiments in which we discarded positive samples from the training set to further illustrate this.
Figure \ref{fig:less_data} shows a graph depicting the results of these experiments run for both our model and Mini-COVIDNet.
While Mini-COVIDNet displayed low performance with 15 and 12 positive examples, before barely outperforming chance with subsequent reduction, our sparse coding model achieved reasonable performance with just 8 examples, obtaining an F1 score of 84.

\subsection{COVID-19}
\begin{table}[]
    \centering
    \begin{tabular}{|c|c|c|}
        \hline
        Model & Macro F1 & Accuracy \\\hline
        VAE + Dense & 16.0 $\pm$ 1.0 & 38\% \\
        Mini-COVIDNet & 64.9 $\pm$ 1.6 & \textbf{75}\% \\
        Our Model & \textbf{67.7 $\pm$ 5.5} & 74\% \\\hline
    \end{tabular}
    \caption{F1 scores (over 3 runs) for the COVID-19 dataset.}
    \label{tab:covid_results}
\end{table}

The BAMC PTX Dataset is not publicly available, therefore we also evaluated our model on the COVID-19 Ultrasound Dataset \cite{born2021accelerating}. This dataset contains 202 LUS videos. Each video has one of four labels: COVID-19, Regular, Bacterial Pneumonia, or Viral Pneumonia.

Table \ref{tab:covid_results} contains the results of our COVID-19 experiments. We reevaluated Mini-COVIDNet to ensure that evaluation remained consistent across models. Our model slightly outperforms Mini-COVIDNet in Macro F1. Mini-COVIDNet achieved an F1 score of 64.9, while our model achieved an F1 score of 67.7. However, the accuracy of Mini-COVIDNet was slightly better than our model on average. We hypothesize that comparable performance between our model and Mini-COVIDNet is observed on this task due to some differences between COVID-19 and PTX detection. Evaluation of the pleural line movement is not necessary for COVID-19 classification; individual frames frequently contain enough information to make a prediction. This lack of emphasis on the pleural line required us to remove a critical part of our pipeline, the YOLOv4 object detection, while the lack of temporal dependencies nullified some of the benefits of 3D sparse coding. Expert knowledge regarding relative feature importance for COVID-19 detection could be used to re-train YOLOv4, allowing us add it back into our system and improve our model's performance on this task.

\subsection{Analyses}

\subsubsection{Sparse Coding Filter Transfer}
\begin{table}[]
    \centering
    \begin{tabular}{|c|c|c|}
        \hline
        Sparse Weights & Macro F1 & Accuracy \\\hline
        Random & 68.0 & 71\% \\
        COVID-19 & 74.7 & 77\% \\
        PTX & \textbf{87.8} & \textbf{88\%} \\\hline
    \end{tabular}
    \caption{Results for the BAMC PTX dataset using 3 different methods for building the sparse filters.}
    \label{tab:transfer_results}
\end{table}
To further quantify the benefits of sparse coding, we evaluated our model on the PTX task using 2 different sets of sparse weights. We borrowed the first set from the COVID-19 sparse coding pre-training. These weights are loosely correlated with the PTX data, as they both came from LUS videos. We randomly generated the second set of weights. Table \ref{tab:transfer_results} contains the results of these experiments and shows the PTX weights outperform the others. Sparse coding can still work with random filters; the filters compete with each other to represent the data and the classifier can still leverage these representations to make its prediction. However, important features that might be specific to our task are lost, such as filters that detect the pleural line or movement.

\begin{figure}[t]
    \centering
    \subfigure[3D Convolutional Sparse Coding Filters]{\label{fig:sparse_filters_a}\includegraphics[width=80mm]{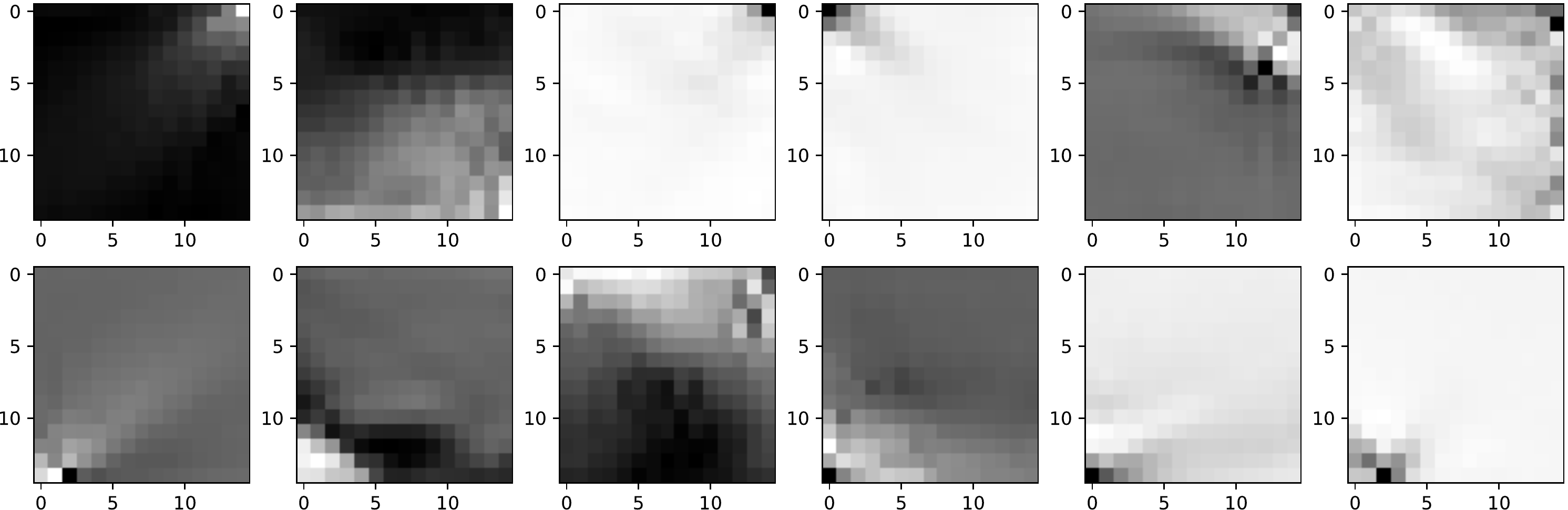}}
    \subfigure[3D Convolutional VAE Filters]{\label{fig:sparse_filters_b}\includegraphics[width=80mm]{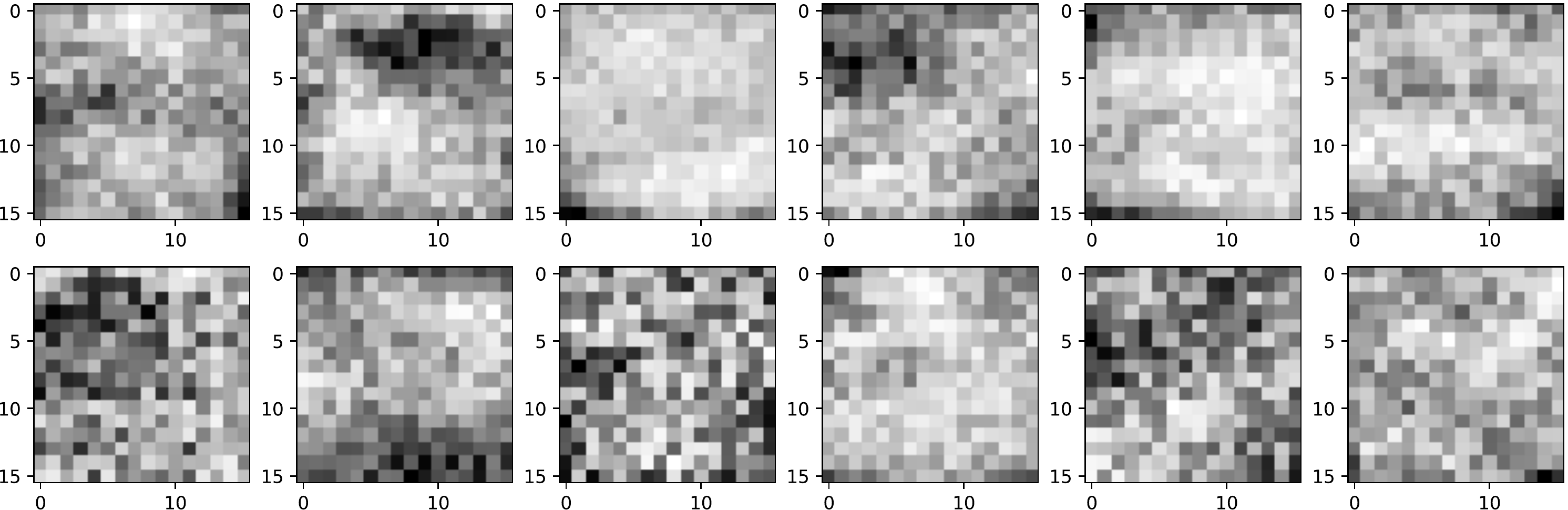}}
    \caption{First frame of the filters learned by our 3D sparse coding model compared with the VAE baseline.}
    \label{fig:sparse_filters}
\end{figure}

\paragraph{Sparse Filter Visualization}
Sparse coding is the backbone of our model. High quality sparse features allow us to learn a lightweight classifier with minimal supervision. Figure \ref{fig:sparse_filters} contains the first frame of our learned five-frame filters compared with the same filters learned by our VAE baseline. The sparse coding filters illustrate Gabor-like structures; some filters are relatively static, capturing edges, while others change drastically frame-to-frame, accounting for movement. A CNN is capable of learning similar structures, however the filters in our VAE are noisy due to the limited training dataset. In contrast, our sparse coding model extracts spatially and temporally relevant features despite being trained with few examples.

\paragraph{Interpretability}
Our attempts to build saliency maps for PTX videos did not produce useful visualizations. We created saliency maps using kernel SHAP \cite{lundberg2017unified}. Our implementation of kernel SHAP differed slightly from standard implementations in that instead of using input instances, we used sparse coding activation maps. SHAP is an additive feature attribution method that models the classifier as a sum of feature contributions. SHAP has a property called local accuracy which dictates that, for each instance, the sum should produce the same value as the classifier. When comparing the classifier’s predictions with SHAP’s values, they were the same only in 92.64\% of the 62 testing videos. We hypothesize that the limited number of training instances may be responsible for not meeting the local accuracy property for all videos. The resultant saliency maps are not consistent and show few differences between correct and incorrect classifications. Hence, they do not provide useful insights for improving the classifier or the end user experience. The YOLO bounding boxes promote interpretability by highlighting the pleural line. This facilitates diagnosis by highlighting a key visual feature for the clinician, and it also indicates which region of the video is fed to the classifier.

\paragraph{iOS Implementation}
\begin{figure}[t]
    \centering
    \includegraphics[width=0.99\columnwidth]{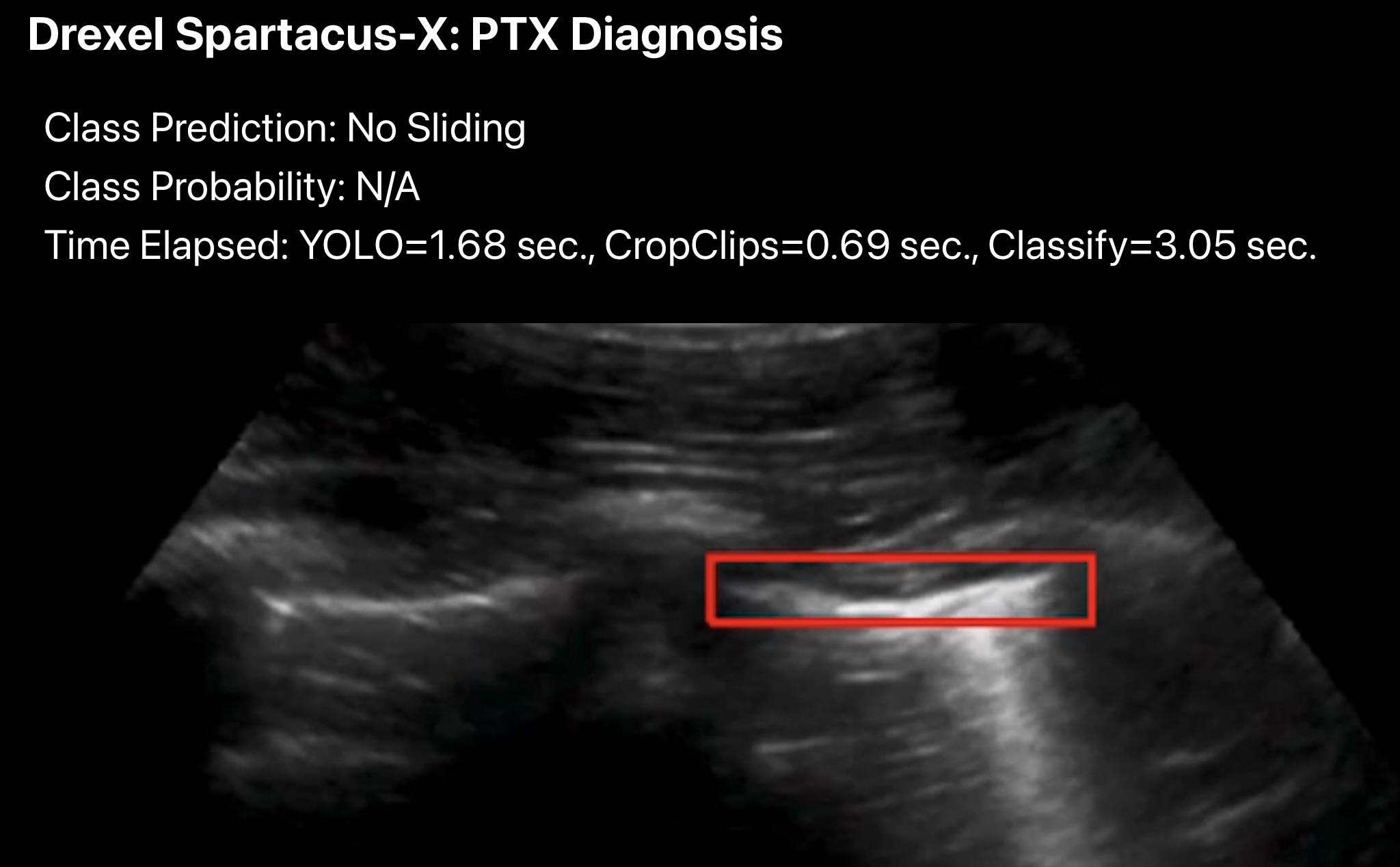}
    \caption{Screenshot of our iOS application.}
    \label{fig:app_screenshot}
\end{figure}
Portability is a focal point of POCUS technology. However, many computer vision techniques have become increasingly computationally expensive. While executing large models on a remote machine might be feasible in some scenarios, having the ability to run on a local device allows for the application to be deployed in even the most austere environments. Therefore, we ensure that our model is capable of running within a reasonable time on a mobile device. For our experiments, we selected a 12.9-inch Apple iPad Pro because it is only 1.54 lbs and has the new M1 processor, a highly efficient and powerful chip that has both a neural engine and a GPU. We built an iOS application using Swift that allows a user to select a set of videos to run through our model. The app replays the LUS video with YOLO bounding boxes around the detected pleural line (Figure \ref{fig:app_screenshot}). These boxes are red (negative) or green (positive) depending on the predicted class at that given point in the video. The app displays the overall prediction and gives the option of exporting a CSV file containing the results. It takes 3.96 seconds (averaged over all test videos) to execute on the iPad: YOLO=0.754 sec, cropping=0.235 sec, classification=2.986 sec. Since the classifier and sparse coding are in the same TFLite file, we do not have individual timing for these components. However, sparse coding accounts for most of this time. During classifier training, we explored both increasing the sparse coding stride and decreasing the number of inner loop updates to improve our execution time and selected optimal hyperparameters while still maintaining close to 90\% accuracy.

\paragraph{Deployment Considerations}
There are a number of considerations that we take into account as we prepare our application for deployment. First, while our application focuses on facilitating diagnosis of a collected video, the collection process itself poses many challenges. In our test cases, experts analyzed the videos to ensure that they were sufficient for PTX differential diagnosis. However, our system may encounter poor quality samples in the field. In the short-term, we can make our model more robust to these cases by augmenting our training data with poor quality samples. We can add these to our present classifier as a third class or feed them to an auxiliary classifier that determines if a video is of sufficient quality for classification. In the long-term, we can integrate our application with other technologies that guide collection \cite{marharjan2020guided}. Another aspect that we must consider is explainability. These models should be used to augment care rather than replace healthcare providers and therefore a great deal of utility lies in the model's ability to report key information to the healthcare provider. We already took a step in this direction by highlighting YOLO regions in our iOS application and by beginning to investigate explainable AI methods. However, in the future we plan to further improve both the model and user interface to maximize the information that healthcare providers receive. We also consider guidelines regarding integrating AI into healthcare outlined by the FDA Digital Health Center of Excellence. Our system does account for some of the guiding principles, such as `leveraging multi-disciplinary expertise' and `tailoring our model design to reflect the intended use of our device.' However, there are some principles that we will need to integrate during deployment. For instance, our system needs to be `directly evaluated in clinical settings, focusing on the performance of a human-AI team, rather than just the model itself.'

\section{Conclusion}
In this work, we presented a LUS video classification system. We focused on developing a model in a constrained setting. Due to the high cost of labeling and collecting ultrasound video, we limited our model to just a few dozen labeled training examples. Therefore, we leveraged expert knowledge to construct a model pipeline that was able to achieve performance on par with human experts on a binary PTX classification task. We provided a robust set of experiments analyzing our performance compared to other architectures, evaluated on two different LUS datasets, and provided a qualitative analysis of the features learned by our sparse coding model. Lastly, we demonstrated that our model was able to run on an iPad Pro in less than 4 seconds and discussed additional deployment considerations that we must address as we proceed with developing our application. Ultimately, we hope our system will increase POCUS adoption and improve quality of care.

\section*{Acknowledgments}
This research was developed with funding from the Defense Advanced Research Projects Agency (DARPA) under the POCUS AI Program (award no. HR00112190076).
The views, opinions and/or findings expressed are those of the author and should not be interpreted as representing the official views or policies of the Department of Defense or the U.S. Government.

\bibliography{aaai23.bib}

\end{document}